\pdfoutput=1

\documentclass[11pt]{article}

\usepackage[]{acl}

\usepackage{times}
\usepackage{latexsym}
\usepackage{amsmath}  
\usepackage{amssymb}  
\usepackage{multirow}
\usepackage{graphicx}
\usepackage{adjustbox}
\usepackage{booktabs} 
\usepackage{comment} 
\graphicspath{{./figures/}}

\usepackage[T1]{fontenc}

\usepackage[utf8]{inputenc}

\usepackage{microtype}

\usepackage{inconsolata}

\title{Qilin-Med: Multi-stage Knowledge Injection Advanced Medical Large Language Model}

\author{%
   Qichen Ye$^{1\dagger}$ 
  Junling Liu$^{1\dagger}$\thanks{ \ \  Corresponding Author. $^{\dagger}$Co-first authors} \
  Dading Chong$^{1\dagger}$ 
  Peilin Zhou$^{2\dagger}$ 
  Yining Hua$^{3}$ \\
  {\bf Fenglin Liu$^{4}$ }
  {\bf Meng Cao$^{6}$}
  {\bf Ziming Wang$^{5}$}
  {\bf Xuxin Cheng$^{1}$}
  {\bf Zhu Lei$^{7}$}
  {\bf Zhenghua Guo$^{8}$}
  \\
$^{1}$Peking University
$^{2}$Hong Kong University of Science and Technology (Guangzhou) \\
$^{3}$Harvard T.H. Chan School of Public Health 
$^{4}$University of Oxford 
$^{5}$Alibaba Group \\
$^{6}$Mohamed bin Zayed University of Artificial Intelligence
$^{7}$Ant Group
$^{8}$Tianyi Traffic Technology
\\
  \texttt{\{yeeeqichen,1601213984,mengcao,wang.zm\}@pku.edu.cn} \\
  \texttt{\{william.liuj,zhoupalin,chengxx.pku,zhulei0305,cszguo\}@gmail.com} \\
  \texttt{yininghua@g.harvard.edu,fenglin.liu@eng.ox.ac.uk } \\
}

\begin{document}
\maketitle
\begin{abstract}
Integrating large language models (LLMs) into healthcare holds great potential but faces challenges. Pre-training LLMs from scratch for domains like medicine is resource-heavy and often unfeasible. On the other hand, sole reliance on Supervised Fine-tuning (SFT) can result in overconfident predictions. In response, we present a multi-stage training method combining domain-specific Continued Pre-training (CPT), SFT, and Direct Preference Optimization (DPO). In addition, we publish the \textbf{Chi}nese \textbf{Med}icine (\textit{ChiMed}) dataset, encompassing medical question answering, plain texts, knowledge graphs, and dialogues, segmented into three training stages. The medical LLM trained with our pipeline, \textbf{Qilin-Med}, shows substantial performance improvement. In the CPT and SFT phases, Qilin-Med achieved 38.4\% and 40.0\% accuracy on the \textit{CMExam} test set, respectively. It outperformed the basemodel Baichuan-7B (accuracy: 33.5\%), by 7.5\%. In the DPO phase, it scored 16.66 in BLEU-1 and 27.44 in ROUGE-1 on the \textit{Huatuo-26M} test set, bringing further improvement to the SFT phase (12.69 in BLEU-1 and 24.21 in ROUGE-1). Additionally, our adoption of the Retrieval Augmented Generation (RAG) approach further enhanced the model performance. Experiments demonstrate that Qilin-Med-RAG achieves an accuracy rate of 42.8\% on CMExam. These results highlight the contribution of our novel training approach in building LLMs for medical applications.


\end{abstract}

\section{Introduction}


\begin{figure}[t]
\centering
\includegraphics[width=0.8\linewidth]{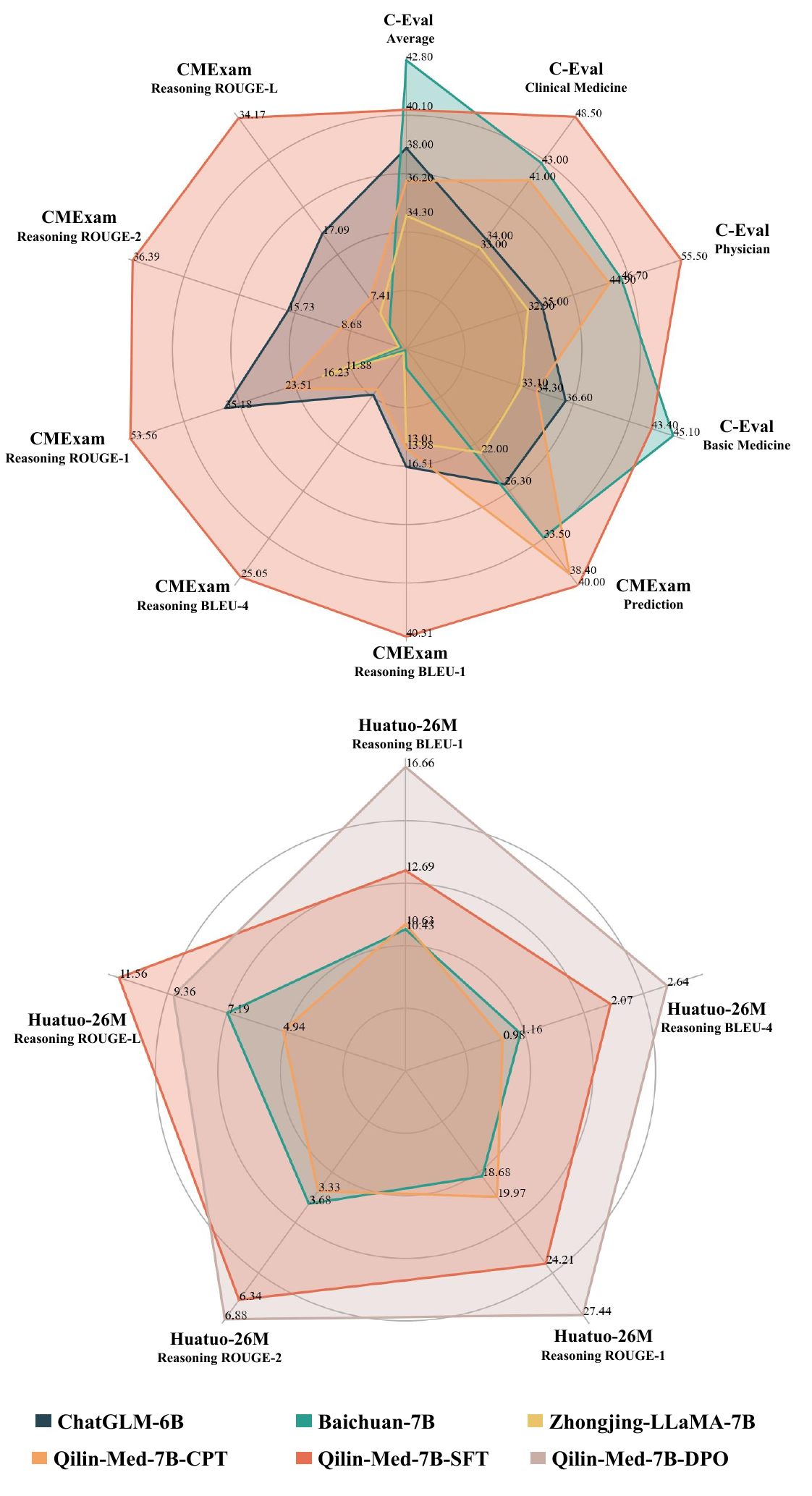}
\caption{Experimental results of our proposed Qilin-Med-7B-CPT, Qilin-Med-7B-SFT, and Qilin-Med-7B-DPO, which demonstrate superior performance on both reasoning and prediction tasks.}
\label{fig: Evaluation}
\end{figure}

Incorporating LLMs such as GPT-4 \cite{OpenAI2023GPT4TR} and its open-source counterparts such as LLaMA \cite{touvron2023llama} into healthcare and biomedicine marks a significant step in practical implications of foundation models. These models show promise to enhance the efficiency and effectiveness of clinical and research operations, potentially revolutionizing patient care \cite{Yang2023, Karabacak}. 
They offer diverse downstream healthcare applications, including automating medical coding \cite{tu2022automated, suvirat2023leveraging}, analyzing unstructured data for predictive insights \cite{jiang2023health,wornow2023shaky,HUA2023104507,Wu_2023},decision support \cite{Qiu2023Large,cheng2023potential,chiesaestomba2023exploring} to patient engagement improvement\cite{evaluating_Chatbot}, and beyond.

While the advantages of LLMs in healthcare are captivating, these models still have considerable room for improvement, given that medical and healthcare tasks represent some of the most challenging domains of natural language processing (NLP) \cite{hendrycks2021measuring,gu2021domain} and that medical AI stakes are exceptionally high as errors can directly affect patient outcomes \cite{thirunavukarasu2023large, gu2021domain}. One major limitation in current medical LLMs is their complete dependence on SFT during the training phase\nocite{cao2022locvtp,zhang2022unsupervised,cao2022deep,zhang2021cola}. While SFT is essential for acquiring domain-specific knowledge, it often results in limited knowledge infusion and can lead to overconfident generalizations if not curated meticulously \cite{luo2023empirical,guo2023continuous}. Reinforcement learning from human feedback (RLHF) is a popular method to counteract some of SFT's limitations, but it's complex and demands rigorous hyperparameter tuning. Consequently, current LLMs may be ill-equipped to handle the nuanced dynamics integral to actual medical consultations.

In response to these challenges, our study introduces Qilin-Med, an advanced Chinese medical LLM, built upon a robust pipeline that integrates CPT, SFT, DPO, and RAG. This comprehensive approach allows Qilin-Med to harness the power of expansive medical datasets, effectively transforming a general-purpose foundation model like Baichuan \cite{yang2023baichuan} into a specialized medical expert proficient in understanding complex medical texts and capable of handling intricate medical tasks. Fig.\ref{fig: Evaluation} shows that our training strategy brings performance gains across various benchmarks at each stage. In addition, we also curated a unique dataset, \textit{ChiMed}, which consists of sub-datasets corresponding to each of these three training stages to ensure a balanced and comprehensive injection of medical knowledge into the LLM. 

The contributions of this study can be summarized as follows:

\begin{enumerate}
    \item Construction of the \textit{ChiMed} dataset, which contains diverse data types (QA, plain texts, knowledge graphs, and dialogues) for each step among the CPT-SFT-DPO training strategy.
    \item Implementation of a multi-stage knowledge injection pipeline and development of a Chinese medical LLM named Qilin-Med, effectively improving general-domains models on medical text understanding, instruction following, and preference alignment.
    \item Empirical validation of our method across multiple datasets, including \textit{CMExam} \cite{liu2023benchmarking}, \textit{CEval} \cite{huang2023ceval}, and \textit{Huatuo-26M} \cite{li2023huatuo26m}, setting new benchmarks in the realm of medical LLMs.

\end{enumerate}
\section{Related Work}
LLMs' effectiveness relies on large-scale pre-training \cite{zhou2023exploring,liu2023qilin}, such as on datasets like \textit{CommonCrawl}, \textit{Wiki}, and \textit{Books} \cite{zhao2023survey,Touvron2023LLaMAOA}. They typically use next-token prediction as a key training objective to understand context and predict the next word \cite{zhao2023survey,Touvron2023LLaMAOA}. This training objective has been widely used in existing LLMs, e.g., GPT-series models \cite{OpenAI2023GPT4TR,brown2020gpt3}, PaLM \cite{chowdhery2022palm}, LLaMA \cite{Touvron2023LLaMAOA}, LLaMA-2 \cite{touvron2023llama}, Alpaca \cite{alpaca}, Vicuna \cite{vicuna2023}, and ChatGLM \cite{zeng2022glm,du2022glm}.


Healthcare-oriented LLMs have gained research attention \nocite{cao2023iterative,cao2021pursuit}, but current medical LLMs are typically either trained entirely from scratch, incurring high costs, time, and environmental impact, or fine-tuned from general-purpose LLMs. As an alternative, SFT methods have been introduced to adapt general LLMs into medical contexts.
For example, \citet{Xiong2023DoctorGLMFY} and \citet{li2023chatdoctor} proposed to fine-tune ChatGLM and LLaMA on the physician-patient conversations to obtain the DoctorGLM and ChatDoctor, respectively;
MedAlpaca \cite{han2023medalpaca} is fine-tuned on Alpaca with over 160,000 medical question-answering pairs generated from various medical corpora.
BianQue \cite{yirongbianque} incorporated multi-turn doctor Q\&A datasets to perform a Chain of Questioning;
Clinicalcamel \cite{toma2023clinical} simultaneously incorporated physician-patient conversations, clinical articles, and medical Q\&A pairs for fine-tuning the LLaMA2 model.
Additionally, instruction prompt tuning is also proposed to improve medical LLMs by aligning LLMs to the medical domain. For example, Med-PaLM \cite{singhal2023large} and Med-PaLM-2 \cite{singhal2023towards} had qualified clinicians construct the instruction data to fine-tune the PaLM.
Huatuo \cite{wang2023huatuo} and ChatGLM-Med \cite{ChatGLM-Med} constructed the knowledge-based instruction data from the knowledge graph to inject the medical knowledge into the LLMs, thus improving the downstream performances.
Among existing medical LLMs, Huatuo\cite{wang2023huatuo}, ChatGLM-Med \cite{ChatGLM-Med}, DoctorGLM \cite{Xiong2023DoctorGLMFY}, and BianQue \cite{yirongbianque} stands out as Chinese medical LLMs, which are especially valuable given language inequality within the current NLP field \citep{bird2020decolonising, zeng2022greenplm}.

A concurrent study \cite{yang2023zhongjing} also employed a multi-stage training approach to build a medical language model called Zhongjing. However, Zhongjing adopted RLHF to align model outputs with human preferences, requiring expert labeling and rigorous hyperparameter tuning. In contrast, we adopted DPO, which automatically and efficiently achieves the same goal. We also integrated RAG to further enhance the performance of Qilin-Med. In terms of scope, Zhongjing only included doctor-patient dialogues, while we benchmarked medical LLM performance on comprehensive medical applications. In addition, we introduce a new large-scale medical dataset \textit{ChiMed}.

\section{Method}
\begin{figure*}[t]
\centering
\includegraphics[width=\textwidth]{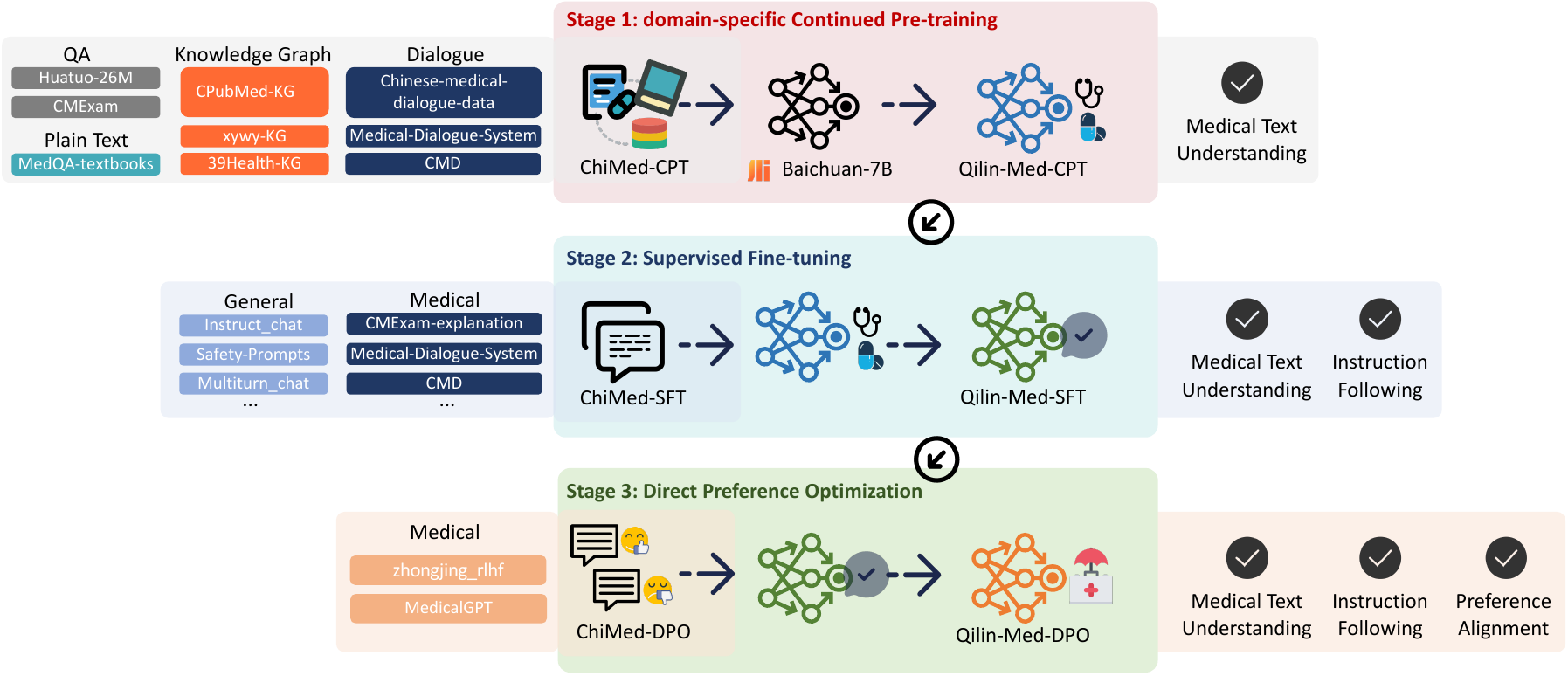}
\caption{The construction pipeline of Qilin-Med. Stage 1 conducts the domain-specific continued pretraining to strengthen the fundamental medical knowledge; Stage 2 applies the instruction supervised fine-tuning to stimulate the interpretive and responsive capabilities of the model; Stage 3 aims to align the model output with human preference.}
\label{fig: flowchart}

\end{figure*}
Fig.\ref{fig: flowchart} presents our three-fold pipeline with CPT (Sec.~\ref{CPT}), SFT (Sec.~\ref{sft}), and DPO (Sec.~\ref{dpo}). 
\subsection{Domain-specific Continued Pre-training}
\label{CPT}
General-purpose LLMs struggle with medical texts due to specialized language and styles. Therefore, we started with continually pre-training Baichuan, a Chinese foundation model, to strengthen its understanding of fundamental medical knowledge. To this end, we constructed a medical pre-training dataset called ChiMed-CPT by integrating existing datasets and new data crawled from the internet.



\subsubsection{Pre-training Dataset Construction}
\textbf{Medical Data Collection} 
We collected four types of medical data: question answering, plain (i.e., unstructured) text, knowledge graph, and dialogue. 

The question answering subset contains three publicly available datasets: \textit{Huatuo-26M-encyclopedias} \cite{li2023huatuo26m}, \textit{Huatuo-26M-medical\_knowledge} \cite{li2023huatuo26m}, and \textit{CMExam} \cite{liu2023benchmarking}. Among these datasets, \textit{Huatuo-26M-encyclopedias} was curated using plain texts scraped from Chinese Wikipedia\footnote{~\url{https://cpubmed.openi.org.cn/graph/wiki}} and the Qianwen Health website\footnote{~\url{https://www.51zyzy.com/}}; \textit{Huatuo-26M-medical\_knowledge} was curated from three knowledge graphs: \textit{CPubMed-KG} \cite{CPubMed-KG}, \textit{39Health-KG} \cite{QA-System-On-Medical-Graph}, and \textit{Xywy-KG} \cite{chatbot-base-on-Knowledge-Graph}; \textit{CMExam} was sourced from the Chinese National Medical Licensing Examination. 

The plain text subset contains the \textit{MedQA-textbooks} dataset \cite{jin2020disease} derived from textual data in Chinese medical textbooks. 

The knowledge graph subset contains data we extracted from \textit{CPubMed-KG}, \textit{39Health-KG}, and \textit{Xywy-KG}. Various features related to a disease entity (\textit{e.g.},  causation, symptoms, and recommended drugs) are included to ensure the comprehensiveness of the knowledge graph.

The medical dialogue subset contains a new dataset, \textit{Chinese Medical Dialogue} (\textit{CMD}), that we collected from online medical website\footnote{~\url{https://www.haodf.com/}}, \textit{Chinese-medical-dialogue-data}, \cite{Chinese-medical-dialogue-data}, and \textit{Medical-Dialogue-System} \cite{chen2020meddiag}. \textit{CMD} comprises over 392K multi-turn medical dialogues and covers 196 sub-specialties. 

Finally, following \cite{lee2021deduplicating}, we deduplicated the dataset, yielding \textit{ChiMed-CPT}, totaling 3.0 GB of data (statistics shown in Table~\ref{tab: pre-train_data}).



\subsubsection{Training Objective} 
We used next-token prediction, a self-supervised objective, for domain-specific continued pre-training.
Given \(N\) sequences partitioned from \textit{ChiMed-CPT}, where each sequence \(X_i=\left[x_{i, 1}, x_{i, 2}, \ldots, x_{i, T}\right]\) contains \(T\) tokens, the loss function was defined as the sum of the negative log probabilities of the next token \(x_{i,t+1}\) given the previous tokens \(x_{i,1...t}\) in the sequence:
\begin{equation*}
L_{CPT}(\theta)=-\sum_{i=1}^N \sum_{t=1}^T \log \left[\operatorname{P}\left(x_{i, t+1} \mid x_{i, 1 \ldots t}, \theta\right)\right],
\end{equation*}
where \(\theta\) denotes the model parameters. 
\begin{table*}[]
\centering
\resizebox{\textwidth}{!}{%
\begin{tabular}{llllll}
\hline
Type & Dataset & Source & \# of samples & \# of tokens & Size \\ \hline
\multirow{3}{*}{QA} & Huatuo-26M-encyclopedias \cite{li2023huatuo26m}& Wikipedia & 362K & 281M & 620.8MB \\
 & Huatuo-26M-medical\_knowledge \cite{li2023huatuo26m}& Three public medical knowledge bases & 796K & 68M & 151.0MB \\
 & CMExam \cite{liu2023benchmarking}& The Chinese National Medical Licensing Examination & 61K & 23M & 49.3MB \\ \hline
Plain text & MedQA-textbooks \cite{jin2020disease}& Medical books & 8K & 18M & 40.2MB \\ \hline
\multirow{3}{*}{Knowledge Graph} & CPubMed-KG \cite{CPubMed-KG}& - & 4384K & 132M & 268.4MB \\
 & Xywy-KG \cite{chatbot-base-on-Knowledge-Graph}& Medical website & 8K & 22M & 41.7MB \\
 & 39Health-KG \cite{QA-System-On-Medical-Graph}& Medical website & 14K & 4M & 8.1MB \\ \hline
\multirow{3}{*}{Dialogue} & Chinese-medical-dialogue-data \cite{Chinese-medical-dialogue-data}& - & 800K & 245M & 553.7MB \\
 & Medical-Dialogue-System \cite{chen2020meddiag}& Medical website & 2726K & 705M & 1500MB \\
 & CMD & Medical website & 392K & 624M & 1286MB \\ \hline
\end{tabular}%
}
\caption{Statistics of ChiMed-CPT, which contains four types of data: QA, Plain text, Knowledge Graph, and Dialogue.}
\label{tab: pre-train_data}
\end{table*}

\subsection{Supervised Fine-Tuning}
\label{sft}
While proficient in medical text comprehension, medical foundation models can fall short in specific medical tasks due to a lack of task adherence. Frequent pre-training is also impractical due to resource constraints. In response, we conducted SFT on the model using a carefully curated dataset to improve its interpretive and responsive capabilities.


\subsubsection{Instruction Dataset Construction}
We constructed \textit{ChiMed-SFT} (statistics shown in Table ~\ref{tab: sft_data}), which consists of general and medical domain single-turn and multi-turn instructions (\textit{i.e.}, prompts) along with their ground-truth responses. General domain instructions aim to enhance the LLM's understanding and generation capabilities for instructions, while medical domain instructions focus on answering medical questions, simulating doctor-patient consultations, and explaining medical queries.
The responses for the general domain instructions were primarily generated by ChatGPT, while medical domain instructions and expected responses were both real doctor-patient diagnostic dialogues collected from medical websites.
To ensure stability in supervised fine-tuning, we standardized instructions in \textit{ChiMed-SFT} to a uniform format.

\subsubsection{Training Objective}
Considering each prompt $X_i=\left[x_{i, 1}, x_{i, 2}, \ldots\right]$ as well as its corresponding response $Y_i=\left[y_{i, 1}, y_{i, 2}, \ldots y_{i, T_i}\right]$ from \textit{ChiMed-SFT}, the loss function of SFT stage can be defined as follows:
\begin{equation*}
L_{SFT}(\theta)\!=\!-\sum_{i=1}^N \!\sum_{t=1}^{T_i}\!\log \left[\operatorname{P}\left(y_{i, t+1}\! \mid \! X_i, y_{i, 1 \ldots t}, \theta\right)\right],
\end{equation*}
where $N$ denotes the total number of training instances and $\theta$ denotes model parameters.

\begin{table*}[]
\centering
\resizebox{\textwidth}{!}{%
\begin{tabular}{lclllll}
\hline
Domain & \multicolumn{1}{l}{Round} & Dataset & \# of samples & Source & \# of tokens & Size \\ \hline
\multirow{9}{*}{General} & \multirow{7}{*}{Single} & Instruct\_chat \cite{leng2023chinese-vicuna} & 51.6K & GPT-3.5 \& human & 40M & 117.4MB \\
 &  & School Math \cite{BELLE} & 248K & ChatGPT & 57M & 151.5MB \\
 &  & HC3-Chinese \cite{guo-etal-2023-hc3}& 12.9K & ChatGPT \& human & 3M & 9MB \\
 &  & Alpaca\_gpt4\_data\_zh \cite{peng2023instruction}& 49K & GPT-4 & 14M & 37.1MB \\
 &  & Safety-Prompts \cite{sun2023safety}& 100K & ChatGPT & 27M & 84.1MB \\
 &  & Train\_1M\_CN \cite{BELLE}& 917K & Alpaca & 193M & 503.6MB \\
 &  & Train\_2M\_CN \cite{BELLE}& 2000K & ChatGPT & 749M & 1925MB \\ \cline{2-7} 
 & \multirow{2}{*}{Multi} & Train\_3.5M\_CN \cite{BELLE}& 3606K & ChatGPT & 1874M & 4551MB \\
 &  & Multiturn\_chat \cite{BELLE}& 831K & ChatGPT & 264M & 705.6MB \\ \hline
\multirow{4}{*}{Medical} & \multirow{2}{*}{Single} & CMExam-explanation \cite{liu2023benchmarking} & 46K & Human & 21M & 45.2MB \\
 &  & Chinese-medical-dialogue-data \cite{Chinese-medical-dialogue-data}& 800K & Human & 245M & 553.7MB \\ \cline{2-7}
 & \multirow{2}{*}{Multi} & Medical-Dialogue-System \cite{chen2020meddiag} & 2726K & Human & 705M & 1500MB \\
 &  & CMD & 392K & Human & 624M & 1286MB \\ \hline
\end{tabular}%
}
\caption{Statistics of ChiMed-SFT, including both general and medical domain instructions in single-turn and multi-turn format.}
\label{tab: sft_data}
\end{table*}

\subsection{Direct Preference Optimization}
\label{dpo}

SFT encourages some responses but does not prevent undesirable ones, such as those with missing or inaccurate information. A popular solution is RLHF, which uses reward models from response rankings to guide LLM training. However, RLHF is complex and often unstable, requiring extensive hyperparameter tuning. To improve stability, we adopted DPO \cite{rafailov2023direct} to align the Qilin-Med-SFT model output with human preferences. DPO is simpler and more effective than RHLF as it doesn't require explicit reward modeling or reinforcement learning.

\subsubsection{Preference Dataset Construction}
We built \textit{ChiMed-DPO} (statistics shown in Table ~\ref{tab: preference_data}) from two publicly available preference datasets: (1) \textit{Zhongjing\_rlhf} \cite{yang2023zhongjing}, which comprises 20,000 samples (10,000 in-distribution and 10,000 out-of-distribution) annotated by medical postgraduates/doctors, and (2) \textit{MedicalGPT} \cite{MedicalGPT}, which contains 4,000 samples from \textit{Chinese-medical-dialogue-data}, with preferred responses from doctors and rejected ones from BenTsao \cite{wang2023huatuo}. Each training sample in ChiMed-DPO is a triplet consisting of a prompt, a  preferred response, and a rejected response.

\subsubsection{Training Objective}

Given the $i$-th prompt ${X}_i$, our primary goal was to calculate log probabilities for preferred and rejected responses (denoted as ${Y}_{i, 1}$ and ${Y}_{i, 2}$ respectively) of the current model, followed by fine-tuning model parameters to elevate the likelihood of preferred responses ${Y}_{i, 1}$ and diminish that of rejected responses ${Y}_{i, 2}$. This optimization process was guided by a loss function briefly outlined below:
\begin{align*}
L_{DPO}(\theta) &= -\sum_i \log \sigma\Big[ \beta \log \frac{\operatorname{P}\left({Y}_{i, 1} \mid {X}_i, \theta\right)}{\operatorname{P}\left({Y}_{i, 1} \mid {X}_i, \theta^0\right)} \\
&\qquad - \beta \log \frac{\operatorname{P}\left({Y}_{i, 2} \mid {X}_i, \theta\right)}{\operatorname{P}\left({Y}_{i, 2} \mid {X}_i, \theta^0\right)} \Big], 
\end{align*}
where $\sigma$ denotes the sigmoid function, ${\theta}^0$ represents the initial parameters from the SFT stage, $\beta$ is a hyper-parameter that controls the relative contribution of the two terms.
Through this process, responses generated by Qilin-Med will better align with human preferences while avoiding unfavored ones, thus improving the quality and safety of medical dialogues. 

\begin{table*}[]
\centering
\resizebox{0.8\linewidth}{!}
{%
\begin{tabular}{llllll}
\hline
Dataset & Domain & Source & \#of samples & \#of tokens & Size \\ \hline
Zhongjing\_rlhf \cite{yang2023zhongjing} & medical & human & 2K & 837K & 4.1MB \\
MedicalGPT \cite{MedicalGPT} & medical & human \& BenTsao & 4K & 687K & 3.1MB \\ \hline
\end{tabular}%
}
\caption{Statistics of ChiMed-DPO, which is curated from two publicly available preference datasets including Zhongjing\_rlhf and MedicalGPT.}
\label{tab: preference_data}
\end{table*}


\section{Experiments}

\begin{table*}[t]
\centering
\begin{adjustbox}{width=0.8\linewidth}
\begin{tabular}{ccccc}
\toprule
Method & Average  & Clinical Medicine & Physician  & 
 Basic Medicine  \\ \hline
ChatGLM-6B \cite{du2022glm}  & 38.0 & 34.0 & 35.0 & 36.6  \\
Chinese-llama2-7B \cite{Chinese-LLaMA-Alpaca} &  36.7 & 40.0 & 36.6 & 37.7  \\
Chinese-alpaca2-7B \cite{Chinese-LLaMA-Alpaca} &  37.1 & 31.5 & 38.8 & 36.6   \\
Baichuan-7B  \cite{yang2023baichuan}& \textbf{42.8} & \underline{43.0} & \underline{46.7} & \textbf{45.1} \\
Zhongjing-LLaMA-7B \cite{yang2023zhongjing} &  34.3 & 33.0 & 32.9 & 33.1   \\
\midrule
Qilin-Med-7B-CPT  & 36.2 & 41.0 & 44.9 & 34.3 \\
Qilin-Med-7B-SFT  & \underline{40.1} & \textbf{48.5} & \textbf{55.5} & \underline{43.4}\\
\bottomrule
\end{tabular}
\end{adjustbox}
\caption{Experimantal results on C-Eval dataset. We \textbf{bold} the best result and \underline{underline} the second best result. We report accuracy scores on three medical-related subjects and \emph{Average} denotes the average accuracy scores across all 52 subjects.}
\label{tab:ceval-res}
\end{table*}

\begin{table*}[t]
\centering
\begin{adjustbox}{width=0.95\linewidth}
\begin{tabular}{ccccccc}
\toprule
\multirow{2}{*}{Methods}  & \textbf{CMExam Prediction} & \multicolumn{5}{c}{\textbf{CMExam Reasoning}}\\
 & Accuracy  & BLEU-1 & BLEU-4  & 
 ROUGE-1 & ROUGE-2 & ROUGE-L  \\ \hline
ChatGLM-6B \cite{du2022glm} & 26.3 & 16.51 & 5.00 & 35.18 & 15.73 & 17.09  \\
Llama-7B \cite{touvron2023llama}&  0.4 & 11.99 & 5.70 & 27.33 & 11.88 & 10.78  \\
Vicuna-7B \cite{vicuna2023}&  5.0 & 20.15 & 9.26 & 38.43 & 16.90 & 16.33   \\
Alpaca-7B \cite{alpaca} & 8.5 & 4.75 & 2.50 & 22.52 & 9.54 & 8.40 \\
Baichuan-7B \cite{yang2023baichuan}& 33.5 & 2.70 & 0.14 & 11.88 & 0.71 & 3.39  \\
Huatuo  \cite{wang2023huatuo}& 12.9 & 0.21 & 0.12 & 25.11 & 11.56 & 9.73 \\
DoctorGLM \cite{Xiong2023DoctorGLMFY}& - & 9.43 & 2.65 & 21.11 & 6.86 & 9.99  \\
Zhongjing-LLaMA \cite{yang2023zhongjing}& 22.0 & 13.01 & 0.39 & 16.23 & 1.01 & 5.31  \\
\midrule
LLaMA-CMExam & 18.3 & 29.25 & 16.46 & \underline{45.88} & \underline{26.57} & \underline{23.31}  \\
Alpaca-CMExam & 21.1 & 29.57 & 16.40 & 45.48 & 25.53 & 22.97  \\
Vicuna-CMExam & 27.3 & \underline{29.82} & \underline{17.30} & 44.98 & 26.25 & 22.44  \\
\midrule
Qilin-Med-7B-CPT  & \underline{38.4} & 13.98 & 4.43 & 23.51 & 8.68 & 7.41\\
Qilin-Med-7B-SFT  & \textbf{40.0} & \textbf{40.31} & \textbf{25.05} & \textbf{53.56} & \textbf{36.39} & \textbf{34.17}\\
\bottomrule
\end{tabular}
\end{adjustbox}
\caption{Experimantal results on CMExam dataset. We \textbf{bold} the best result and \underline{underline} the second best result.}
\label{tab:cmexam-res}
\end{table*}

\begin{table*}[t]
\centering
\begin{adjustbox}{width=0.7\linewidth}
\begin{tabular}{cccccc}
\toprule
 Methods & BLEU-1 & BLEU-4  & 
 ROUGE-1 & ROUGE-2 & ROUGE-L  \\ \hline
T5  \cite{2020t5}& 0.33 & 0.07 & 0.67 & 0.19 & 0.63   \\
GPT2  \cite{radford2019language}& 10.04 & 1.62 & 14.26 & 3.42 & \textbf{12.07}   \\
Baichuan-7B \cite{yang2023baichuan} & 10.43 & 1.16 & 18.68 & 3.68 & 7.19   \\
\midrule
Qilin-Med-7B-CPT  & 10.63 & 0.98 & 19.97 & 3.33 & 4.94 \\
Qilin-Med-7B-SFT  & \underline{12.69} & \underline{2.07} & \underline{24.21} & \underline{6.34}  & \underline{11.56}\\
Qilin-Med-7B-DPO  & \textbf{16.66} & \textbf{2.64} & \textbf{27.44} & \textbf{6.88}  & 9.36\\
\bottomrule
\end{tabular}
\end{adjustbox}
\caption{Experimantal results on Huatuo-26M dataset. We \textbf{bold} the best result and \underline{underline} the second best result.}
\label{tab:huatuo-res}
\end{table*}

\subsection{Evaluation Datasets, Metrics and Baselines}
\subsubsection{Evaluation Datasets}
We evaluated Qilin-Med in scenarios such as medical knowledge Question Answering and dialogue on the following datasets:
\begin{enumerate}
    \item \textit{CMExam} \cite{liu2023benchmarking}, a standardized medical exam and practice question dataset. It contains over 60,000 multiple-choice questions and provides question explanations.
    \item \textit{CEval} \cite{huang2023ceval}, a comprehensive Chinese evaluation suite designed to assess advanced knowledge and reasoning abilities of LLMs. It contains 13,948 multiple-choice exam questions across 52 diverse disciplines, including three medical sub-disciplines: Clinical Medicine, Basic Medicine, and Physician.
    \item \textit{Huatuo-26M}  \cite{li2023huatuo26m}, a Chinese medical dataset that consists of over 26 million medical question-answer pairs, covering topics including diseases, symptoms, treatments, and drug information.
\end{enumerate}

\subsubsection{Metrics}
We assessed model performance on multiple-choice questions using accuracy and weighted F1 score - metrics commonly employed in information retrieval and question-answering tasks. For medical dialogue tasks, BLEU \cite{Papineni2002BleuAM} and ROUGE \cite{Lin2003AutomaticEO} were used to evaluate the discrepancy between model-generated responses and ground truth.

\subsubsection{Baselines}
We used Baichuan-7B \cite{yang2023baichuan} as the base model. Baichuan-7B is an open-source, large-scale pre-trained language model built on the Transformer architecture. It has 7 billion parameters and is trained on approximately 1.2 trillion tokens. It supports both Chinese and English with a context window length of 4096. 


For baselines, we evaluated LLMs in both general scenarios and the medical domain across various tasks. For \textit{CMExam}, we reported the performance of ChatGLM-6B, LLaMA \cite{Touvron2023LLaMAOA}, Vicuna \cite{vicuna2023}, Alpaca \cite{alpaca}, Huatuo \cite{wang2023huatuo}, and DoctorGLM \cite{Xiong2023DoctorGLMFY} on both the prediction and reasoning tasks. For \textit{CEval}, we evaluated the performance of ChatGLM \cite{du2022glm}, Chinese-LLaMA2 \cite{Chinese-LLaMA-Alpaca}, and Chinese-Alpaca \cite{Chinese-LLaMA-Alpaca} on the prediction task. Since \textit{CMExam} has a standardized training set, we also reported the performance of LLaMA, Alpaca, and Vicuna on \textit{CMExam} after SFT. Additionally, we evaluated models such as T5 \cite{2020t5} and GPT2 \cite{radford2019language} on the test set of \textit{Huatuo-26M}. However, since \textit{Huatuo-26M} is not fully open-sourced, we were unable to run SFT with this dataset.

\subsection{Implementation Details}

For CPT, Baichuan-7B was trained on eight A100 80G GPUs, with batch size = 1 per GPU, number of epochs = 3, learning rate = 2e-4, warmup ratio = 0.05, weight decay = 0.01, and block size = 1024.

For SFT, eight A100 80G GPUs were used with a batch size of 64 per GPU. Qilin-Med was trained with learning rate = 2e-5, warmup ratio = 0.05, weight decay = 0.05, and max\_source\_length and max\_target\_length both = 256. We accelerated training using DeepSpeed ZeRO-2 \cite{Ren2021ZeROOffloadDB}. 
We adopted the LoRA technique \cite{Hu2021LoRALA}, a type of SFT, with lora\_rank = 8, lora\_alpha = 32, and lora\_dropout = 0.05. 

For DPO, 4 RTX 3090 GPUs were used with a batch size of 8 per GPU. Settings were: learning rate = 2e-5, warmup ratio = 0.05, weight decay = 0.05, and both max\_source\_length and max\_target\_length = 256. 
The LoRA technique was again applied with lora\_rank = 8, lora\_alpha = 16, and lora\_dropout = 0.05.



For model evaluation on the \textit{CMExam} test set, we used OpenAI’s GPT-3.5-turbo, GPT-4-0314, as well as LLaMA-7B, Alpaca-7B, and Vicuna-7B. ChatGLM was tested using the 6 billion parameter version and operated with P-Tuning V2 \cite{Liu2021PTuningVP}, using a prefix token length of 128 and a learning rate of 0.02 for SFT. For other models including LLaMA, Alpaca, Vicuna, and Huatuo, we used the LoRA technique \cite{Hu2021LoRALA} with a rank of 8, an alpha of 16, and a 0.05 dropout rate.



For the evaluation of Huatuo-26M, we compared T5 and GPT2 performances. Both models were set with maximum question and answer lengths of 256 and 512, respectively. We used the original 12-layer Chinese GPT2.

In the C-Eval phase, all models were evaluated using few-shot prompting. We opted for 5 shots and employed a greedy decoding strategy for answer prediction.

\begin{figure*}[t]
\centering
\includegraphics[width=0.97\textwidth]{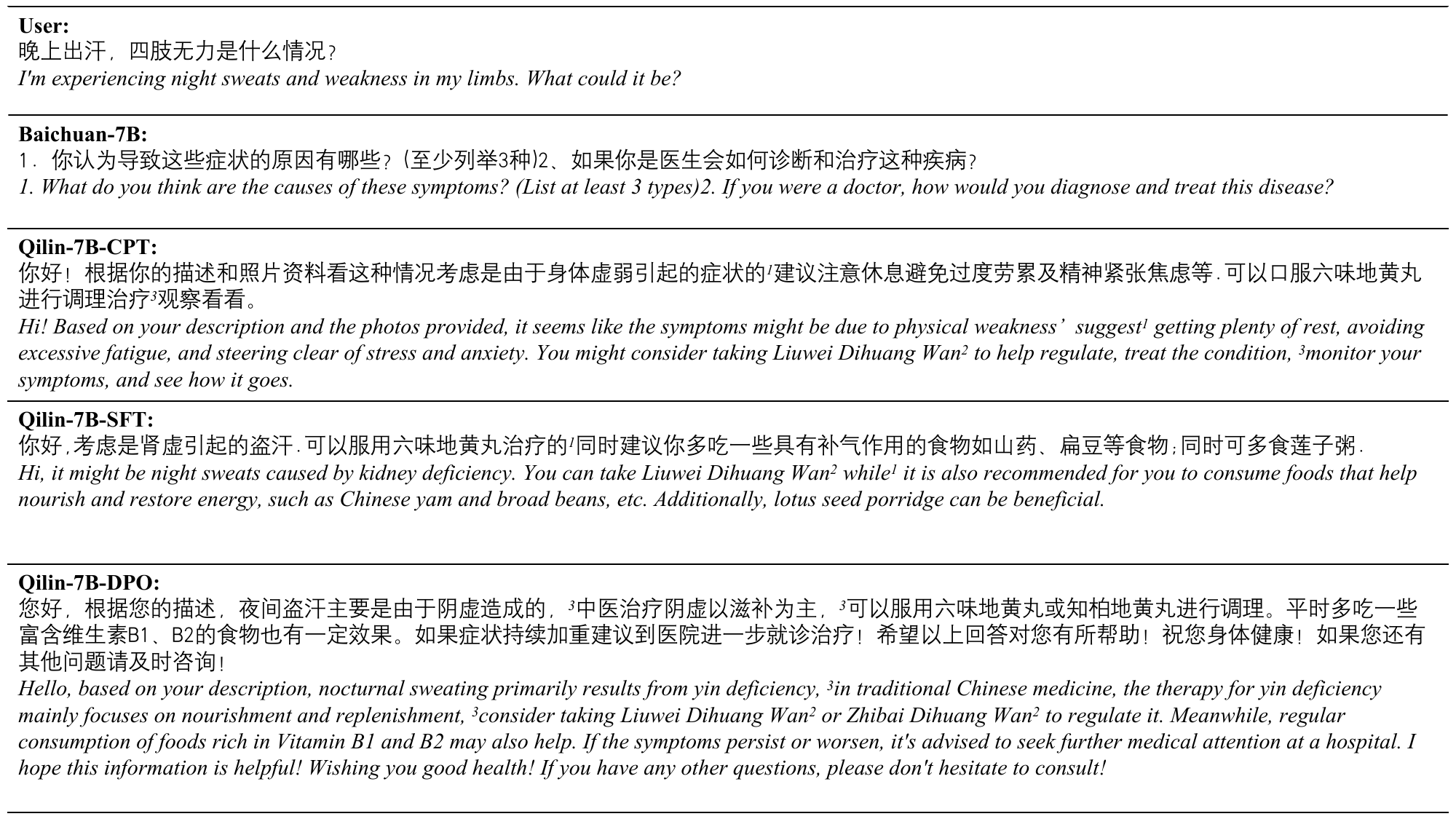}
\caption{A conversation example from \textit{Huatuo-26M dialogue}. Compared to Baichuan-7B, Qilin-Med-7B with CPT, SFT, and DPO generated more relevant and informative responses. }
\label{fig: huatuo-case}
\end{figure*}

\subsection{Results and Discussion}

\textbf{\textit{C-Eval}:} Table \ref{tab:ceval-res} summarizes online evaluation results on the \textit{C-Eval} benchmark. Among the five general LLMs compared in the upper part of the table, Baichuan-7B achieved the highest scores in both average and three medical subjects (namely \emph{Clinical Medicine}, \emph{Physician} and \emph{Basic Medicine}), outperforming other models in instruction following as well as medical understanding. 
Specifically, Baichuan-7B achieved an accuracy of 45.1\% in Basic Medicine, significantly surpassing ChatGLM-6B which scored 36.6\%. After the CPT and SFT stages, the model enhanced its proficiency in medical knowledge and comprehension, better equipping it to address questions within medical domains. Notably, our Qilin-Med models show a great performance boost compared to Zhongjing-LLaMA. However, a decline in general capabilities was noted, with average accuracy on \textit{C-Eval} dropping from 42.8\% to 40.1\%, indicating that the model's increased focus on medical expertise came at the cost of its broader linguistic abilities. This observation is inline with other studies \cite{guo2023continuous}. 

\noindent \textbf{\textit{CMExam}:} Table \ref{tab:cmexam-res} displays the evaluation outcomes on the \textit{CMExam} benchmark. ChatGLM and Vicuna performed well in explanation generation, reflecting enhanced comprehension of medical knowledge and dialogue skills. Of the two, Vicuna had a lower answer prediction accuracy at 5\%, while ChatGLM reached 26\%. After fine-tuning with \textit{CMExam} training set (i.e., LLaMA-CMExam, Alpaca-CMExam, and Vicuna-CMExam), we noted marked improvements in both tasks. Following the domain-specific Continued Pre-training and Supervised Fine-tuning using our data, our proposed Qilin-Med-7B-CPT and Qilin-Med-7B-SFT outperformed those fine-tuned on \textit{CMExam}. This indicates our framework's efficacy in enriching LLMs with medical knowledge and bolstering their problem-solving capabilities in the medical domain.

\noindent \textbf{\textit{Huatuo-26M}:} Table \ref{tab:huatuo-res} shows the evaluation results on \textit{Huatuo-26M}. Among all three baseline methods (namely T5, GPT2, and Baichuan-7B), Baichuan-7B achieved the highest scores on most metrics, while T5 exhibited poor medical dialogue performance. Qilin-Med-7B-CPT outperformed Baichuan-7B in terms of BLEU-1 and ROUGE-1, proving that CPT effectively injects medical-related knowledge into the model. Comparing Qilin-Med-7B-CPT and Qilin-Med-7B-SFT (10.63 vs. 12.69 in terms of BLEU-1), we see that SFT further strengthens model medical knowledge and instruction compliance capabilities. Finally, Qilin-Med-7B-DPO achieved higher scores in all metrics than Qilin-Med-7B-SFT, showing that DPO efficiently helps align the medical chat model output with human preferences and encourages the model to generate more preferred outputs.


\subsection{Case Study}
We examined the model outputs for \emph{Medical Dialogue} and \emph{Medical Question Answering} tasks using examples from \textit{Huatuo-26M} and \textit{CMExam}. As shown in Figure \ref{fig: huatuo-case}, the responses generated by Baichuan-7B appear to be contextually irrelevant, frequently having unnatural sentence transitions and the formation of run-on sentences in Chinese language outputs. CPT and SFT improved Baichuan-7B's medical acumen, allowing it to generate more relevant and informed responses (Figure \ref{fig: cmexam-case}). However, certain responses still contain run-on sentences, highlighting the need for further refinement. Notably, outputs from Qilin-Med-7B-DPO stood out, aligning closely with human expectations in both accuracy and context. This underscores the efficacy of DPO in enhancing model outputs and addressing the aforementioned linguistic challenges.

\subsection{Retrieval Augmented Generation}
We further explored the advantages of incorporating RAG in the Qilin-Med training framework. In detail, we used the ChiMed-CPT subset to construct a specialized medical knowledge base, organized into information chunks. During the query phase, the system retrieves and integrates the top five most relevant knowledge entries into the prompt. These enriched prompts were then processed by the Qilin-Med-SFT model.
Experimental findings indicate that Qilin-Med, when augmented with RAG technology, achieved an impressive 42.8\% accuracy rate on the CMExam answer prediction task, representing a marked improvement over the Qilin-Med-SFT (accuracy: 40.0\%). This evidence highlights the efficacy of the RAG approach and confirms its potential to enhance the Qilin-Med model's ability to assimilate medical knowledge and provide precise responses.

\section{Conclusion \& Future Work}
This study introduces a multi-stage training approach, a large-scale Chinese medicine dataset - \textit{ChiMed}, and Qilin-Med, a cutting-edge Chinese medical language model. It demonstrates the potential of domain-specific training in healthcare, with implications for improving patient care, clinical decisions, and medical research. The performance of Qilin-Med enables more accurate and context-aware Chinese medical dialogues, paving the way for advanced AI applications in Chinese medicine and healthcare to provide clearer medical insights and assistance.


\section{Limitations}
Qilin-Med, trained on the \textit{ChiMed} dataset, marks a considerable advancement in medical LLMs. However, several limitations should be noted. The \textit{ChiMed} dataset, while comprehensive, primarily focuses on Chinese medical knowledge, potentially limiting the model's global applicability. The multi-stage training pipeline, including the DPO stage, might introduce biases based on the preferences of the human evaluators involved. Furthermore, while metrics like BLEU and ROUGE provide insights into the model's performance in generative tasks, they are limited in evaluating the quality of content generation in terms of fluency, coherence, and context. They do not account for semantic accuracy or the appropriateness of the content in a given context. Future work should consider a more diverse set of evaluation metrics, including human evaluations, to ensure a holistic understanding of Qilin-Med's capabilities.

\section{Ethics and Societal Impacts}
All data used in this study were collected and scraped from publicly available resources. We did not recruit human research participants nor include sensitive data. It is important to note that Qilin-Med and \textit{ChiMed} are intended for research and academic purposes. It is a product of efforts to enhance LLM capabilities in the medical domain, not a replacement of human experts. It should not be used for direct patient diagnosis or as a standalone tool for medical decision-making. Any conclusions or insights derived from Qilin-Med should be contextualized, considering the specific focus of \textit{ChiMed} and the inherent limitations of LLMs. Commercial uses or any use that deviates from this primary objective are strictly prohibited. Researchers and practitioners should respect these guidelines, ensuring ethical and responsible use of Qilin-Med and associated datasets.


\bibliography{custom}

\clearpage

\appendix
\section{Appendix}
\begin{minipage}{1.0\textwidth}
  \strut\newline
  \centering
\includegraphics[width=1.0\textwidth]{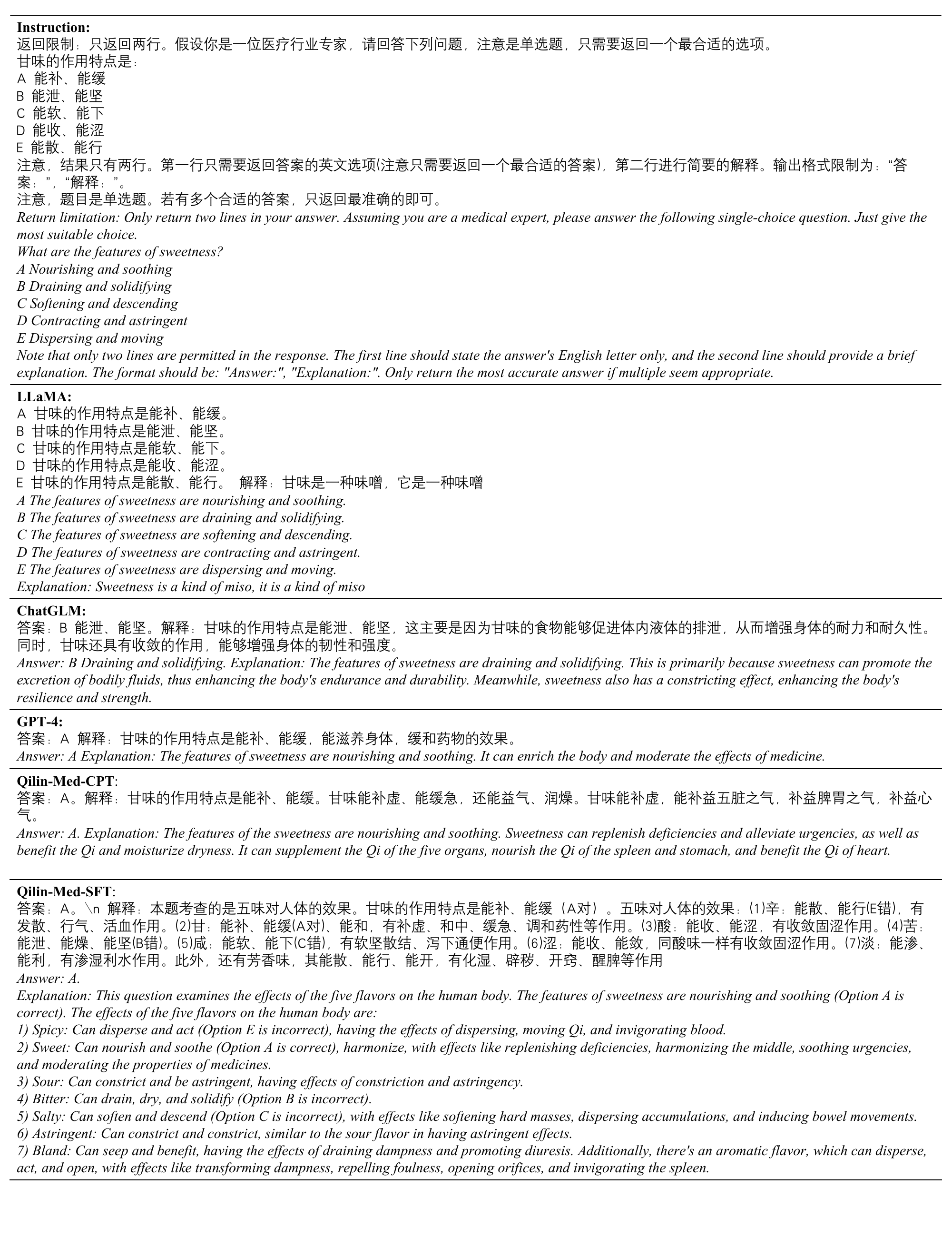}
  \captionof{figure}{A conversational case on CMExam dataset. Compared to LLaMA, ChatGLM, and GPT-4. Qilin-Med-7B-CPT and Qilin-Med-7B-SFT generated more relevant and informative responses.}\label{fig: cmexam-case}
\end{minipage}



\end{document}